\pdfoutput=1
\documentclass[journal]{IEEEtran}
\ifCLASSINFOpdf
\else
\fi

\usepackage{amsmath}
\usepackage{array}

\usepackage{stfloats}
\usepackage{float}
\usepackage{graphicx}
\usepackage{wrapfig}
\usepackage[bookmarks=false]{hyperref}
\usepackage{xcolor}
\hypersetup{
    colorlinks,
    linkcolor={blue!50!black},
    citecolor={blue!50!black},
    urlcolor={blue!80!black}
}
\usepackage[utf8]{inputenc} % swedish chars

\usepackage{natbib}

\usepackage{amsmath}
\usepackage{graphicx}
\usepackage[colorinlistoftodos]{todonotes}
\usepackage{amssymb}
\usepackage{bm}
\usepackage{amsthm}

\usepackage{tabularx}
\usepackage{amsfonts}
\usepackage[figurename=Fig.,labelfont=bf]{caption}
\usepackage{pgfplots}

\newtheorem{assumption}{Assumption}

\begin{document}
\title{Imitation Learning with Concurrent\\ Actions in 3D Games}

\author{Jack~Harmer$^1$,
        Linus~Gisslén$^{*1}$,
        Jorge~del~Val$^{*1}$,
        Henrik~Holst$^1$,
        Joakim~Bergdahl$^1$,
        Tom~Olsson$^2$,
        Kristoffer~Sjöö$^1$,
        Magnus~Nordin$^1$,% <-this % stops a space
\thanks{Correspondance to jharmer@ea.com}% <-this % stops a space
\thanks{$^1$Electronic Arts, SEED, Stockholm, Sweden}% <-this % stops a space
\thanks{$^2$Electronic Arts, DICE, Stockholm, Sweden}% <-this % stops a space
\thanks{$^*$These authors contributed equally}}% <-this % stops a space
%% \thanks{Manuscript received March, 2018}}

%% \markboth{Journal of \LaTeX\ Class Files,~Vol.~14, No.~8, August~2015}%
%% {Shell \MakeLowercase{\textit{et al.}}: Bare Demo of IEEEtran.cls for IEEE Journals}



\maketitle

\begin{abstract}
In this work we describe a novel deep reinforcement learning architecture that allows multiple actions to be selected at every time-step in an efficient manner. Multi-action policies allow complex behaviours to be learnt that would otherwise be hard to achieve when using single action selection techniques. We use both imitation learning and temporal difference (TD) reinforcement learning (RL) to provide a 4x improvement in training time and 2.5x improvement in performance over single action selection TD RL. We demonstrate the capabilities of this network using a complex in-house 3D game. Mimicking the behavior of the expert teacher significantly improves world state exploration and allows the agents vision system to be trained more rapidly than TD RL alone. This initial training technique kick-starts TD learning and the agent quickly learns to surpass the capabilities of the expert.
\end{abstract}

\begin{IEEEkeywords}
deep reinforcement learning, imitation learning
\end{IEEEkeywords}

% \fi
\IEEEpeerreviewmaketitle

\section{Introduction}
\IEEEPARstart{I}{n} recent years the field of reinforcement learning (RL) \citep{Sutton1998} has undergone a renascence, due to the transformative powers of deep neural network architectures. A key strength of these architectures is their ability to be used as arbitrary function approximators. This ability has allowed neural network based model free RL techniques to solve a number of challenging tasks that were previously intractable. The techniques described by \cite{Mnih2015b} demonstrate the power of this, in their work they develop an algorithm (DQN) that employs a neural network to estimate the value of high dimensional input states and use a bootstrapping technique to train it. Put simply, this algorithm minimizes the difference between the networks estimate of the value of the current state and that of a target value, where the target value is simply the networks predicted value of the next state plus any rewards that were received in-between the two states. The target value is more grounded in reality than the initial guess because it is partly made up of rewards gained through an agents interactions with the environment, and as such they show that these updates allow the network to learn an accurate value function. They further demonstrate the capability of this algorithm by training agents to play Atari 2600 video games using high dimensional raw pixel values as the input state. Due to the success of this work and a renewed interest in the field, a large number of improvements to this algorithm have now been suggested in the literature, including but not limited to: A modification that reduces the bias of the value function estimate (DDQN) \citep{VanHasselt2015}, a technique for improving the data efficiency of the algorithm, by adding a type of prioritisation to the experience replay memory sampling scheme \citep{Schaul2015}, adding noise to specific layers to improve exploration \citep{Fortunato}, breaking the action value function into two components, one that models the value of the state and one that models the per-action advantage  \citep{Wang2016}, and also modeling the state value function as a distribution \citep{Bellemare}.

Unlike these previous off-policy techniques, \cite{Mnih2016} propose an algorithm that moves away from an experience replay based training regime. Instead, they describe an architecture (A3C) that performs updates using data from a large number of simultaneously running agents. They show that training using multiple agents, each with their own version of the environment, decorrelates updates in a similar manner to memory sampling in DQN, with the added benefit of improved exploration, training speed and stability.

Despite these advances, algorithms based around temporal difference (TD) RL are computationally expensive and can take a significant amount of time to train. Training using TD RL is only effective if the target value is more grounded in reality than the current estimate. This condition is only satisfied when there is a net accumulation of reward between states, otherwise training simply updates one guess towards another. Thus training using TD RL is particularly problematic in reward sparse environments where it might require many specific consecutive actions to receive a reward. Consider a racing car game where the reward is scaled inversely with lap time and received after completing a lap. This task is extremely difficult to solve when using TD RL. To receive a reward and thus perform one useful update, an agent would have to select the correct action for many thousands of steps at a stage when the agent has no understanding of the world. 

The technique known as reward shaping \citep{Ng1999} can alleviate some of the problems with reward sparse environments. When using this technique the reward function of the task is changed by an expert, who understands the objective, in order to encourage behaviours that help the agent solve the task. However, great care has to be taken because it is not always trivial to tweak the rewards without significantly altering the nature of the task at hand. It is also often difficult to break down a complicated task into a number of smaller sub-tasks amenable to shaping. 

Tasks where large action spaces are required are also difficult to train when using TD RL, because the probability of selecting the correct action, in order to receive a reward, decreases as the size of the action space increases. Further, credit assignment also becomes more problematic \citep{Sutton1998}.

When training agents to interact in complex environments with large action spaces, the behaviour associated with having a single action per time step (SAPS) policy, as is almost always the case in RL, is often undesirable. For example, running forward whilst strafing and shooting in a video game is an effective strategy that is impossible to achieve when using SAPS architectures. When solving problems where multiple actions per time step are required, most networks architectures rely on modeling all possible combinations of actions as separate distinct actions \citep{Mnih2015b, Wang2016b}. However when using large actions spaces, the dramatic increase in the number of possible action combinations severely limits the applicability of such techniques. For example, a typical modern video game controller might have around 20 controls, modeling all possible combinations of these inputs would require a policy which outputs $\sim 10^6$ (\(2^{20}\)) probabilities. Joint action representations in such a large action space make it much harder for the agent to learn the value of each of the true actions, and do not take advantage of the underlying relationships between different sets of individual actions. For these reasons, an algorithm that allows multiple output actions per time step (MAPS) should improve the performance of such an agent. 

A powerful technique that can be used to speed up training is to teach by example. The idea being that instead of using a domain expert to break down rewards into more fine-grained rewards, an expert can be used to demonstrate the desired behaviour. Then, the network is left to determine how to change its policy in order to match the expert behaviour. This technique is known as imitation learning (IL) \citep{Subramanian2016,Hester2017a,Le2018,Andersen2018,Nair2017,Zhang2018,Gao2018}. Imitation learning provides the agent with prior knowledge about effective strategies for behaving in the world. Combining TD RL with IL allows an agent to learn from it’s own experiences, and helps to avoid situations where the skill of an agent is limited by the skill of the teacher. Learning via imitation can either be the goal itself, or an auxiliary task that is used to help achieve another goal by bootstrapping off the behaviour of an expert

\section{Contributions}
Motivated by the goal of adding neural network controlled AI agents to future games, in order to increase levels of player immersion and entertainment, we describe a technique for training an agent to play a 3D FPS style game. In comparison to 2D games such as those on the Atari 2600 platform, 3D FPS games are a particularly challenging problem for RL. This is mainly due to the factors described previously but also because of the partially observed nature of such games, and the challenges related to exploring large state spaces. We develop an in house FPS game using a modern game engine in order to test the performance of agents in scenarios with high visual fidelity graphics. 

We present an A3C derivative algorithm that combines supervised imitation learning (learning via guidance from an expert teacher) with temporal difference RL (learning via trial and error), throughout training; using only a small amount of expert data. Imitation learning is used as an auxiliary task in order to help the agent achieve it's primary goal, playing the game.

We describe a neural network architecture that outputs multiple discrete actions per time step without having to model combinations of actions as separate actions, and describe a loss function that allows the policy to be trained.  Combining multi-action per time step RL with imitation learning in this manner allows higher quality expert data to be used, as it circumvents the difficulties associated with recording expert data when the expert is limited to single action per time step interactions with the environment. We call the resulting model Multi-Action per time step Imitation Learning (MAIL).
\begin{itemize}
\item We present a neural network architecture that outputs multiple discrete actions per time step (MAPS), as well as a loss function for training the multi-action policy.
\item We describe a technique for training this algorithm using a combination of imitation learning and temporal difference reinforcement learning (MAIL).
\item We describe how these techniques can be used to teach an agent to play a challenging fully 3D first person shooter (FPS) style video game, an important milestone on the way to training neural networks to play modern AAA FPS games.
\end{itemize}
%\IEEEpubidadjcol

\section{Related Work}
Training using IL is by definition off-policy and as such is typically limited to off-policy training techniques \citep{Hester2017}, which tend to be less stable than on-policy algorithms \citep{Mnih2016}, or to being carried out as a pre-training step \citep{Silver2016,Schaal1997}. Indeed in one of the earliest examples of this \cite{Schaal1997} use imitation learning as a pretraining step during which they train the value function and policy, after which they continue to train the network using reinforcement learning alone. However, they found that even for simple nonlinear tasks task-level imitation based on direct-policy/value learning, augmented with subsequent self-learning, did not provide significant improvements to learning speed over pure trial-and-error learning without demonstration. When imitation learning is used as a pre-training step, the policy is also often prone to collapse due to the limited state-space coverage of the expert data; that is, models tend to over-fit to the data instead of learning a general solution. This can be mitigated by using a large amount of expert training data, as in the work by \cite{Silver2016}. However, the time, effort and cost associated with collecting such data is often a limiting factor in the effective deployment of these techniques. \cite{Ross2010} develop a technique called DAGGER that iteratively generates new policies based on polling the expert policy. However, DAGGER requires additional feedback during training and therefor requires the expert to be available, as such it is impractical for long training runs or when access to the expert is limited.  Another example of work in this category, i.e. combining expert data and reinforcement learning, is Approximate Policy Iteration with Demonstration (APID) \citep{Kim2013}. Here an expert's trajectories are used to define linear constraints which are used in the optimization made by a Policy Iteration algorithm.

\cite{Wang2016b} look at the problem of performing multiple actions per time step in a Q-learning setting by modeling combinations of actions as separate actions and estimating the value for each of these. However as previously mentioned this technique is problematic when using large actions spaces, due to the rapid increase in the number of possible action combinations that have to be accounted for. Moreover, their results are only contextualized in Least Squares Policy Iteration (LSPI) and TD learning with linear value function approximation, rather than an actor critic deep reinforcement learning approach. \cite{Sharma2017} attempt to reduce the combinatorial explosion by reducing the action space into a number of sub-actions that are mutually exclusive for a given scenario, such as the actions representing left and right in a video game. However, this requires specific prior knowledge of the action space and only partly offsets this effect. Furthermore, they still have to evaluate each possible combination separately to select a joint action, which is not efficient for large action spaces. \cite{Lillicrap2016} describe a technique for selecting multiple actions per time step in a continuous action setting, however such techniques are notoriously difficult to train due to their brittleness and hyperparameter sensitivity \citep{Haarnoja2018}. 

\section{Preliminaries} \label{section:preliminaries}

We consider a Markov Decision Process (MDP) described by the tuple $\langle \mathcal{S}, \mathcal{A}, \mathcal{P}, r \rangle$, where $\mathcal{S}$ is a finite set of states, $\mathcal{A}$ is a finite set of actions containing $N$ possible actions, $\mathcal{P}: \mathcal{S}\times \mathcal{A} \times \mathcal{S} \rightarrow [0,1]$ is the transition probability kernel and $r: \mathcal{S} \rightarrow \mathbb{R}$ is the reward function. In the standard SAPS context the action space is:

\begin{equation}
\mathcal{A}_{sa} = \left\{A_1, A_2, ..., A_N \right\} .
\end{equation}

Consequently, the action $a_t\in \mathcal{A}_{sa}$ at each timestep $t$ is limited to be one of the available actions. Each action is selected using a stochastic policy $\pi: \mathcal{S}\times \mathcal{A} \rightarrow [0,1]$, so that $\mathbf{a}_t \sim \pi(a_t|s_t)$. When using high-dimensional state spaces, these policies are typically parametrized by a deep neural network of weights $\theta$; that is, $p(a_t|s_t) = \pi_\theta(a_t|s_t)$.

In a MAPS setting, we allow the agent to select multiple actions at each timestep. Hence, the action space becomes the space of all possible subsets of different elements of $\mathcal{A}_{sa}$. We can easily see we can represent those combinations with binary vectors of dimension $N$, i.e., elements of $\mathbb{Z}_2^{N}$, where each component $a_i$ indicates whether the action $A_i\in\mathcal{A}_{sa}$ was taken or not:

\begin{equation}
\mathcal{A}_{ma} = \left\{ (a_1, a_2, ..., a_{N}) | a_i\in \{0,1\} \right\} .
\end{equation}

Here, we have added the zero action to indicate that none of the possible actions was taken. Clearly under this framework $\mathcal{A}_{sa}\subset \mathcal{A}_{ma}$, since the SAPS space could just be defined as $\mathcal{A}_{sa}=\{a\in \mathcal{A}_{ma} | \sum_{i=1}^N a_i =1\}$. We denote the marginal probability of a single component $a_i$ with $p(a_i)$.

We let $R_t = \sum_{k=0}^\infty \gamma^k r_{t+k}$ denote the total cumulative discounted reward. We also let $V^\pi(s)=\mathbb{E}\left[R_t|s_t=s\right]$ denote the state value function, $Q^\pi(s,a) = \mathbb{E}\left[ R_t | s_t = s, a_t = a\right]$ denote the action value function, and $A^\pi(s,a) = Q^\pi(s,a) - V^\pi(s)$ denote the advantage function corresponding to policy $\pi$.

In value-based deep reinforcement learning methods, the action value function is approximated by a deep neural network of parameters $\phi$, this is, $Q^\pi(s, a) \approx Q^\pi_\phi(s, a)$. In some cases it is a direct estimate of the value function that is approximated $V^\pi(s) \approx V^\pi_\phi(s)$. The parameters of the policy are updated to minimize the loss over each batch of experiences to approximately satisfy Bellman's equation:
\begin{equation}
L_v(\phi) = \frac{1}{K}\sum_{k = 1}^K \| r_k + \gamma V^\pi_{\phi'}(s'_{k}) - V^\pi_\phi(s_k)\|^2 ,
\end{equation}
where $\phi'$ denotes the parameters of a separate \emph{target network}, which are clamped in the loss function, and $k$ indexes each experience $(s_k,a_k,r_k,s'_k)$ on a batch containing $K$ experiences. In policy-based methods, by contrast, it is the policy network $\pi_\theta$ that is updated following estimates of the policy gradient
\begin{equation}
\nabla_\theta \mathbb{E}[R_t] = \mathbb{E} \left[ \sum_{\tau} \nabla_\theta \log\pi_\theta(a_{\tau}|s_{\tau})\Psi^\pi_{\tau}  \right] ,
\end{equation}
where the expectation is taken accross the set of all possible paths and $\Psi^\pi_{\tau}$ can be a variety of choices, among which the advantage function $A^\pi_\tau$ is one of the most common, since it yields almost the lowest possible variance of the gradient estimator \citep{Schulman2015}. In actor-critic frameworks, the advantage function is estimated using a value network as in value-based methods; this is, $A_{t} = \hat{R}_t - V^\pi_\phi(s_t)$, where $\hat{R}_t$ is a TD estimate of $R_t$. Hence, the estimate of the policy gradient over a set of $M$ independent rollouts $B_1,..., B_M$ is:
\begin{equation}
\nabla_\theta \mathbb{E}[R_t] \approx \frac{1}{M}\sum_{i=1}^M \left( \sum_{t \in B_i} \nabla_\theta \log\pi_\theta(a_t| s_t)A_t \right) .
\end{equation}
%
%where $\eta_\tau = -\nabla_\theta \log\pi_\theta(a_\tau| s_\tau)A_\tau $.

%\section{Deep Reinforcement Learning in big combinatorial spaces}
\section{Algorithm}

In a discrete multi-action setting it can easily be seen that the cardinality of the space grows exponentially with the number of available actions as $O\left(2^{N}\right)$. As a consequence, a policy which models the the probability of all possible set of event becomes intractable as $N$ becomes large. In order to circumvent this problem, in this paper we make the following structural assumption for the policy:

\begin{assumption} Under the policy $\pi$, each component $a_i$ of $a \in \mathcal{A}_{ma}$ is conditionally independent given the state $s$. That is
\label{mainassumption}
%\begin{equation}
%%p(a|s)=p(a_1,a_2,...,a_N|s)=\prod_{i=1}^N p(a_i|s)
%p(a_i, a_j|s)=p(a_i|s)p(a_j|s) \quad i \ne j, \forall i,j \in \{1, ..., N\} ,
%\end{equation}
%or
\begin{equation}
p\left( a | s \right) = \prod_{i=1}^{N} p(a_i|s) .
\end{equation}
\end{assumption}
This assumption simplifies the problem to the one of modeling $N$ parameters instead of $2^N$, making it tractable, at expense of losing representation power and all conditional dependencies. In theory this can be restrictive, since a high probability of a joint action would necessarily imply high marginal probabilities of single actions, and vice versa. However, we find that in practice the flexibility of this family of policies is enough to outperform single-action policies. The relaxation of this assumption and the exploration of the tradeoff between flexibility and performance is left for future work.

We hence model our policy as a set of Bernoulli random variables whose probabilities $p(a_i)=\phi_{i;\theta}(s_t)\in[0,1]$ are outputs of a deep neural network. This is
\begin{equation} 
\pi_\theta(a_t|s_t) = \prod_{i=1}^N a_{i;t}\phi_{i;\theta}(s_t) + (1-a_{i;t})(1-\phi_{i;\theta}(s_t)) , \label{modelprob}
\end{equation}
where $a_{i;t}$ denotes the $i_{th}$ component of $a_t$. To sample from this distribution we just have to sample each action independently, relying on assumption \ref{mainassumption}.

As discussed in section \ref{section:preliminaries}, an estimate of the policy gradient is
\begin{equation}
%\eta_t = -\nabla_\theta \log \pi_\theta(a_t|s_t) A_t \label{genericupdate}
\hat{g} = \frac{1}{M}\sum_{i=1}^M \left( \sum_{t \in B_i} \nabla_\theta \log \pi_\theta(a_t|s_t) A_t \right) , \label{genericupdate}
\end{equation}
where $A_t$ is an estimate of the advantage function at time $t$. In our case, we can easily particularize this expression following equation (\ref{modelprob}):
\begin{equation}
\begin{split}
& \log \pi_\theta(a_t|s_t) \\
& = \log \left(\prod_{i=1}^N a_{i;t}\phi_{i;\theta}(s_t) + (1-a_{i;t})(1-\phi_{i;\theta}(s_t)) \right) \\
& = \sum_{i=1}^N \log \left(a_{i;t}\phi_{i;\theta}(s_t) + (1-a_{i;t})(1-\phi_{i;\theta}(s_t)) \right) \\
& = \sum_{i=1}^N a_{i;t} \log(\phi_{i;\theta}(s_t)) + \vphantom{\sum_i^N}  (1-a_{i;t})\log(1-\phi_{i;\theta}(s_t)) \\
& = -H(a_t, \phi_\theta(s_t)) , \label{modelupdate}
\end{split}
\end{equation}
where for convenience we use $H$ to denote the standard cross-entropy formula; although we remark that it is not cross-entropy in an information-theoretic framework, since $a_t$ is not sampled from a categorical distribution. Our proposed gradient update for rollouts $B_1,...,B_M$ is thus
\begin{equation}
\hat{g} = - \frac{1}{M}\sum_{i=1}^M \left( \sum_{t \in B_i} \nabla_\theta H(a_t, \phi_\theta(s_t)) A_t \right) . \label{genericupdate}
\end{equation}

\subsection{Imitation Learning}
TD RL can be highly inefficient when training agents to perform tasks in complex environments with sparse reward and/or high dimensional action spaces. A powerful yet simple technique for improving pure TD learning is to train the network to imitate the behaviour of an expert in the domain, be it another algorithm or a human expert. \cite{Silver2016} describe an effective technique for this and manage to train neural network controlled agents to play the game of Go to superhuman performance levels. They perform imitation learning as a pre-training step before RL. They sample from a large repository of expert human data (30 million examples) and use the data to train a deep neural network to maximise the likelihood of selecting the expert action, given the same input. 

One of the major problems associated with pre-training with imitation learning, is over-fitting to the expert data. The network remembers exactly what actions to perform for a specific input image in the expert training data set, instead of learning a robust and general solution to the problem. Then, when new states are encountered during TD learning, the agent is incapable of selecting an action intelligently. \cite{Silver2016}  work around this by training using a very large expert data set and are helped by the fully observed nature of the task. 

Due to the difficulties involved in collecting a large amount of expert data, we take a different approach. Instead of applying imitation learning as a pre-training step, we apply it at the same time as TD RL as a way of regularizing the TD learning. Each batch update is comprised of both expert and live agent data. At every update step, the network predicts the action of the expert, from a sample of the expert data, whilst learning a policy that maximises the discounted future reward of the live agent stream. Training the network in this way allows the network to maintain a valid TD learning compatible state, throughout training. 

To encourage generalisation, we add Gaussian noise to the expert data inputs and apply dropout after every layer, except the outputs. Dropout is only used for the expert data. To prevent the final performance of the agent from being limited by the quality of the expert data, the IL loss weighting factor, \(\lambda_{E}\), is linearly decayed from the start of training.

We found that training the value function using the expert data reduced the performance and stability of the agent. As such the expert data was only used to train the policy whereas the value function was trained using pure TD RL alone.

Our final policy update for the MAIL network for a set of $M$ independent rollouts of live experiences $B_1, ..., B_M$ and $M$ independent batches of expert data $B^E_1, ..., B^E_M$ is thus:

\small
\begin{align*}
\hat{g}  = - \frac{1}{M}\sum_{i=1}^M \nabla_\theta \left(\sum_{t \in B_i}  \vphantom{\sum_{e\in B^E_i}}H(a_t, \phi_\theta(s_t)) A_t \right. \\
 + \left. \lambda_{E} \sum_{e\in B^E_i} H\left(a_e, \phi_\theta(s_e)\right) \right) .
\end{align*}
\normalsize

\section{Experimental Methods}

\begin{figure*}
\centering
\includegraphics[width=0.9\linewidth]{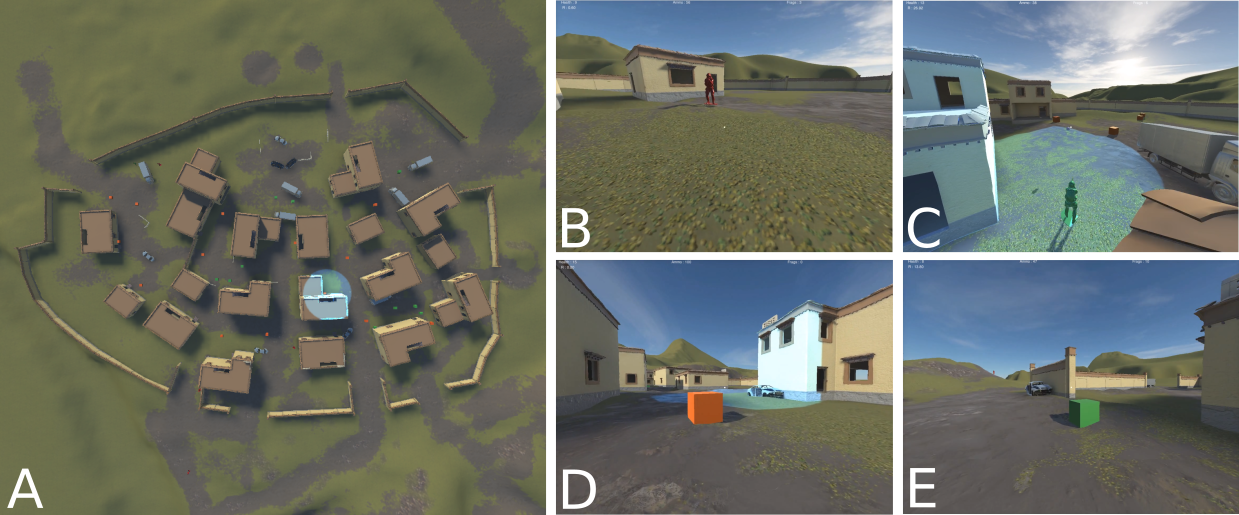}
\caption{\label{fig:org05ff803}
The environment. A: Top down overview of the play area. B: Enemy. C: Region-of-interest (light blue region) and agent. D: Health. E: Ammo}
\end{figure*}

All agents were trained using a batched version of the A3C algorithm (A2C), similar in design to \cite{Babaeizadeh2016}, with the addition of the modifications described previously.  

The performance of the following algorithms were evaluated:
\begin{itemize}
\setlength\itemsep{0.5em}
\item SAPS with TD learning
\item MAPS with TD learning
\item MAIL without TD learning
\item MAIL with \(\lambda_{E}\) decay over 15M steps
\item MAIL with \(\lambda_{E}\) decay over 50M steps
\end{itemize}
In the MAIL experiments, each training batch consisted of approximately 50\% on-policy live agent data and 50\% expert data. Salient information for the expert data is provided in Table \ref{tab:ep_info}. Expert data was generated prior to training by recording episodes of human play. At each time-step, the following information was stored in a memory buffer: input observation, expert action vector, reward, terminal state and game features vector. The game features vector contained the agents health and ammo to simulate the on-screen text that a human player can read.  

\begin{table}[htbp]
\renewcommand{\arraystretch}{1.3}
\caption{\label{tab:ep_info}
Expert data statistics}
\centering
\begin{tabular}{lr}
\hline
Observations & \(\sim 40000\)\\
Episodes & 30\\
Mean score & 47\\
Score std & 33\\
\hline
\end{tabular}
\end{table}

An in-house developed 3D FPS video game was used as the training environment. In the game, rewards are received for eliminating enemies, collecting health and ammo and for finding and occupying a region-of-interest on the map. The location of the health, ammo boxes and region-of-interest change at regular intervals throughout each episode to a random location. Enemies spawn in waves and navigate towards the agent, attacking once within range. Figure \ref{fig:org05ff803} provides a visual overview of the environment and demonstrates the key features of the game.

\begin{figure}[htbp]
\centering
\includegraphics[width=9cm]{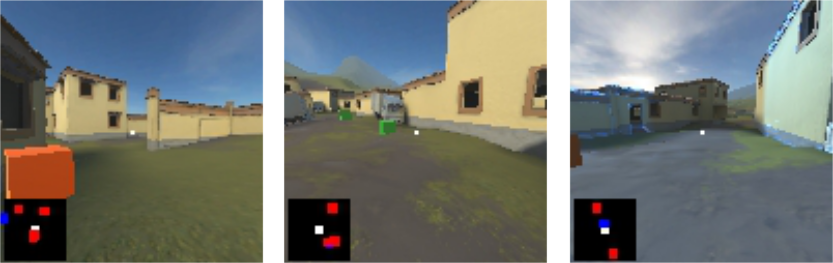}
\caption{\label{fig:orgaf7b8b1}
Example input observations. Left: The agent can see a red health box and some buildings in the main view. The agent can also see a number of red enemies and the blue region-of-interest marker in the radar view. Centre: An example of a green ammo box. Right: The agent has reached the region of interest, indicated by blue lighting on the floor around the agent.}
\end{figure}

At each time step, the agent observes a 128x128 pixel RGB image (see Figure \ref{fig:orgaf7b8b1}) of the agents first-person view. A small short range radar is visible in the bottom left corner of the agents input image. The agent is also provided with a game features vector that contains information related to the agents health and ammo. Experiments indicated that using 128x128 RGB image observations improved the agents performance relative to 84x84 observations, due to the high visual fidelity of the environment. 

The range of actions that the agent can perform include 13 distinct actions that control: translation \((x, y, z)\), head tilt, rotation (multiple torque settings), firing, no-op (SAPS tests).
%% \begin{itemize}
%% \setlength\itemsep{0.5em}
%% \item Translation \((x, y, z)\)
%% \item Head tilt
%% \item Rotation (multiple torque settings)
%% \item Firing
%% \item No-op (SAPS tests)
%% \end{itemize}
In the MAPS experiments, any combination of the actions in the action set can be selected at every step.

\begin{table}[htbp]
\renewcommand{\arraystretch}{1.3}
\caption{\label{tab:org3b19564}
Network Architecture}
\centering
\begin{tabular}{lrl}
\hline
Layer & N & Details\\
\hline
Conv. 1 & 32 & 5x5 kernel, stride 2\\
Conv. 2 & 32 & 3x3 kernel, stride 2\\
Conv. 3 & 64 & 3x3 kernel, stride 2\\
Conv. 4 & 64 & 3x3 kernel, stride 1\\
Linear & 256 + 2 & 2 input features\\
LSTM & 256 & \\
Policy & 13 & \\
Value & 1 & \\
\hline
\end{tabular}
\end{table}

\begin{figure}[htbp]
\centering
\includegraphics[width=6cm]{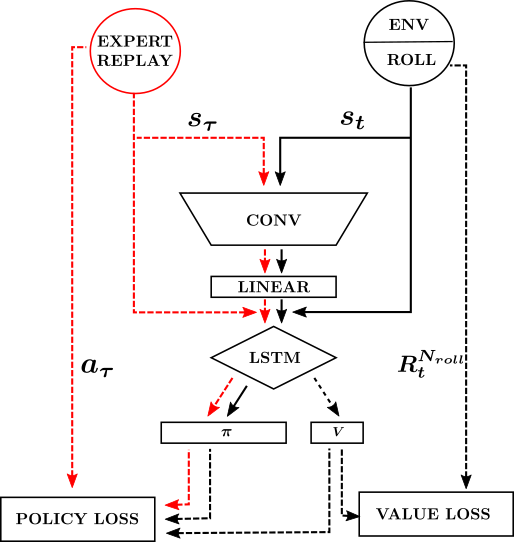}
\caption{\label{fig:orgd200b42}
MAIL Neural network architecture. Solid black lines represent the flow of data during inference. Dashed lines represent the flow of data during training. Red lines represent the flow of expert data.}
\end{figure}

We used the base network architecture that is shown in Table \ref{tab:org3b19564} for all experiments. The high level input features (ammo and health) were concatenated to the output of the linear layer, prior to the LSTM (see Figure \ref{fig:orgd200b42}). The inputs were normalised by their maximum possible value. Training parameters that were global to all experiments are shown in Table \ref{tab:org8eef730}.

For the IL experiments, Gaussian noise was added to both the input observations (mean 0, std 0.1) and high level features vector (mean 0, std 0.3) Dropout was applied to all hidden and convolutional layers. We used dropout values of 60\% and 50\% for the convolution and hidden layers respectively. Dropout was not applied when processing live agent data. Dropout was chosen over \(L_{2}\) weight regularisation, to reduce the risk of the network finding non-optimal local minima \citep{IanGoodfellowYoshuaBengio2016}, instead of more general solutions with larger weights. For the experiments using IL decay, the expert prediction loss factor, \(\lambda_E\), was linearly decayed from 1.0 to 0.0 over the number of decay steps for the experiment.

\begin{table}[htbp] %!t
\renewcommand{\arraystretch}{1.3}
\caption{Global parameters}
\label{tab:org8eef730}
\centering
\begin{tabular}{lr}
\hline
Image size & [128,128,3]\\
Input features size & 2\\
Batch-size & 80\\
Roll-out length & 20\\
Gradient norm clipping & 0.5\\
Optimiser & Adam\\
Initial learning rate & $$1e-4$$\\
Final learning rate & $$1e-5$$\\
training steps & $$75e6$$\\
\(\lambda_{E}\) & 1\\
\hline
\end{tabular}
\end{table}

The main results are shown in Figure \ref{fig:org775a600}. SAPS A3C (red curve) reaches a final score of \(\sim40\). MAPS A3C (blue curve), reaches a final score of \(\sim25\). MAIL (green curve) reaches a final score of \(\sim100\).

When using RL alone MAPS A3C is more difficult to train than SAPS A3C due to the difficulties associated with credit assignment when training using multiple actions (see Introduction). The main problem of training using a SAPS agent however, is that the policy imposes a hard limit on the maximum capability of the agent. This capability is lower than that of an optimal MAPS agent because SAPS policies are a subset of MAPS policies. Indeed, in the best case scenario, a very simple environment where there is no advantage associated with carrying out multiple actions simultaneously, this capability can at best only match that of a MAPS agent. However, the relatively high update frequency of the agent (\(\sim15\) actions per second) offsets some of the problems associated with single action per time step updates in this game. Running forward whilst strafing can, to a limited extent, be approximated by selecting the forward action in one frame and then the strafe action in the next.

During the early stages of training, the MAPS agent trains more rapidly than the SAPS agent. In the SAPS agent case, firing limits its opportunity to move which in turn adversely affects its ability to pick up boxes and get to the region-of-interest. In the MAPS case, because firing has no effect on locomotion, and allows the agent to hit enemy targets, the agent quickly learns that firing is generally a positive action. However, this initial advantage disappears halfway through training, at which point the SAPS agent learns the benefits of interleaving fire actions and locomotion actions (see red vs blue line in Figure \ref{fig:org775a600}). The performance of the SAPS agent eventually surpasses that of the MAPS agent since it is less affected by credit assignment issues. 

\section{Analysis}
\begin{figure}[htbp]
\centering
\includegraphics[width=9.0cm]{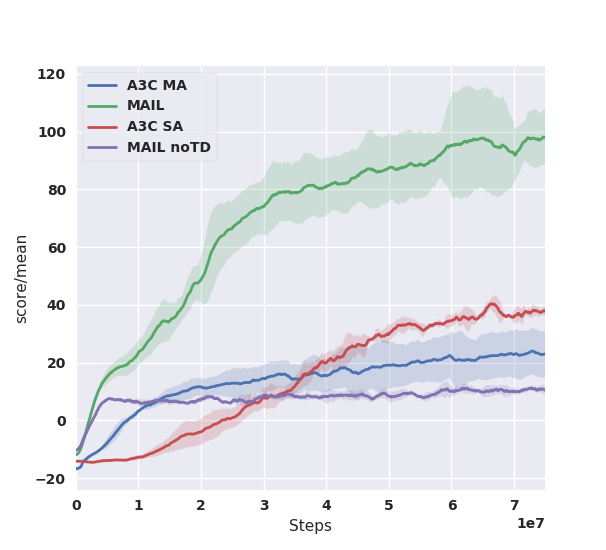}
\caption{\label{fig:org775a600}
Multi-action A3C, MAPS (blue). MAIL (green). MAIL without TD RL (purple). Single Action A3C, SAPS (red). The mean and standard deviation over 5 runs are shown for each result.}
\end{figure}

In this environment, the absolute magnitude of the theoretical performance difference between MAPS A3C and SAPS A3C is difficult to determine because, due to the difficulties of training using TD RL, the SAPS and MAPS agents never reach an optimal policy (see Introduction). 

MAIL significantly outperforms both SAPS A3C and MAPS A3C, reaching a final score \(\sim 2.5\) x higher than SAPS A3C and 4x higher than MAPS A3C. It allows an effective policy to be learnt in far fewer steps than when using TD RL alone, exceeding the final score of SAPS A3C after just 17.5M steps, a \(\sim 4\) x reduction in training time (see \url{https://www.youtube.com/watch?v=LW20UbquVBU} for example agent behaviour).

This speed-up is most pronounced in the early stages of training when reward sparsity severely limits the effectiveness of TD learning updates; imitation learning provides useful feedback at every training step from the very start of training. Supervised learning allows the vision system to be trained much more rapidly than TD RL. Further, mimicking the behaviour of the expert significantly improves the exploration of state-space in comparison to the unguided random actions in the early stages of TD RL. The MAIL agent quickly learns to collect boxes whilst heading towards the region-of-interest; this behaviour can be seen after less than one hour of training ($<1$M steps). From the point of view of the agent, this rapid increase in agent capability significantly reduces reward sparsity and kick-starts the next phase of training, in which temporal difference learning dominates. During this final phase the agent learns to surpass the capabilities of the expert. The mean score of the expert human player was \(\sim 47\); significantly lower than the final score of the MAIL agent, but significantly higher than the other algorithms. 

The trained MAIL agent takes full advantage of the MAPS architecture, and typically performs between 1 and 4 actions at once, learning behaviors such as running forward whilst simultaneously moving sideways, turning and shooting. The MAIL agent performs a similar number of actions per step as the expert teacher taking full advantage of the ability to perform multiple actions concurrently. The concurrent action architecture proved critical for effective imitation learning as it was not possible to record high quality expert human data when limiting the expert to performing single actions at a time, in this game.

\begin{figure}[htbp]
\centering
\includegraphics[width=9.0cm]{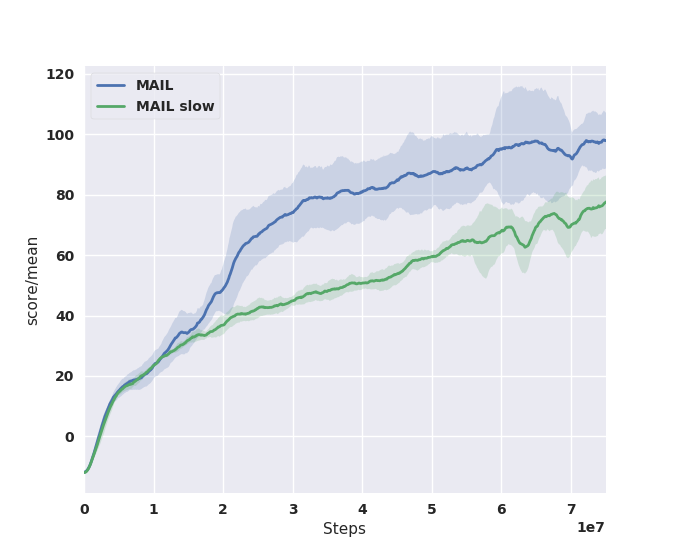}
\caption{\label{fig:org6b3f852}
MAIL (15M) vs MAIL slow decay (50M)}
\end{figure}

%% \begin{figure*}[!t]
%% \centering
%% \includegraphics[width=14.5cm]{behaviours2.png}
%% \caption{Example behaviours. Yellow: seeking region of interest, while picking up boxes. Blue: Patrolling the region-of-interest. Red: Seeking Ammo}
%% \label{fig:orge599d98}
%% \end{figure*}

To better understand how TD RL and IL affect the final MAIL agents capability, we also trained a network without using TD RL (purple curve in Figure \ref{fig:org775a600}). The MAPS IL-only network achieved a final score of \(\sim 15\) , significantly lower than all other training runs. This score was achieved after just 5M steps, with no further improvement during the remaining 70M training steps.  The results show that, when combined with IL, TD RL has a positive contribution in the very early stages of training; after \(\sim\) 2M steps the performance of MAIL surpasses that of pure IL MAIL. At \(\sim\) 13M steps the MAIL agents score is twice that of a pure IL agent. It appears that by forcing the network to learn a solution that maximises future reward, TD RL also helps the agent find a more general solution, which allows it to extract more useful information from the expert data; however, testing this hypothesis is left for future work. To asses whether the expert data eventually starts to limit the performance of the agent we compare the performance of a MAIL agent using two different decay rates for the expert data loss (Figure \ref{fig:org6b3f852}). The run using IL data with a higher decay rate reaches a higher final score, suggesting that IL eventually holds back the performance of the agent. These results also seem to indicate that IL learning reduces the variance in agent performance across games, which can be seen in Figure \ref{fig:org6b3f852}.

\begin{figure}[htbp]
\centering
\includegraphics[width=8.5cm]{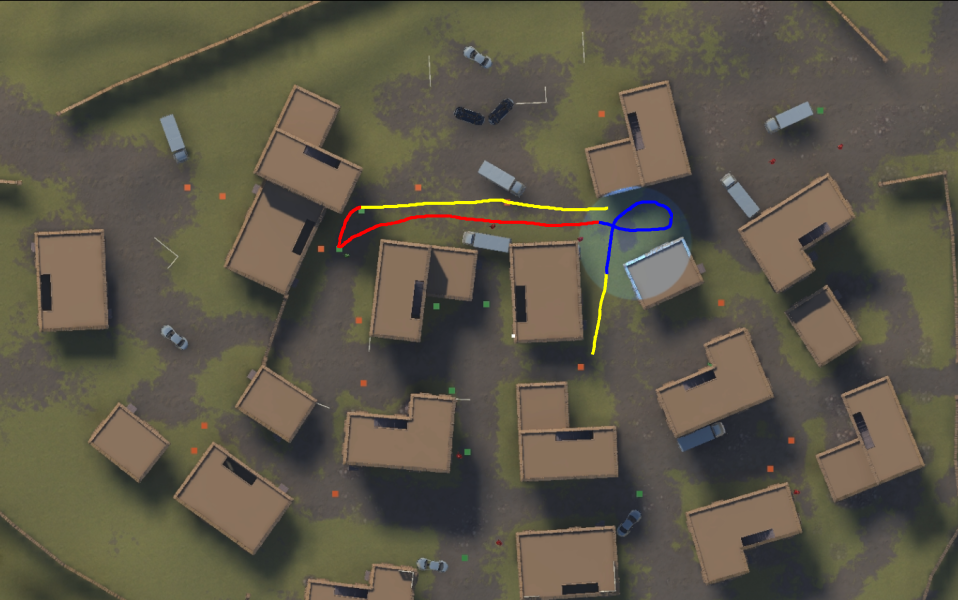}
\caption{Example behaviours. Yellow: seeking region of interest, while picking up boxes. Blue: Patrolling the region-of-interest. Red: Seeking Ammo}
\label{fig:orge599d98}
\end{figure}

Interestingly, the behaviour of the trained MAIL agent is distinctly modal in nature. Its behaviour changes significantly depending upon the agents current state. Certain triggers, such as the agent running low on ammo, cause the agent to drastically alter its style of play. These advanced sub-behaviors arise naturally without deliberately partitioning the network to encourage them, i.e. without using concepts such as manager networks. With even more efficient training techniques, deeper networks with simple architectures might be capable of much higher level reasoning than is currently observed. Examples of some of the observed behaviours of the agent include:  searching for the waypoint, searching for ammo/health, patrolling the region-of-interest, attacking enemies, fleeing enemies due to low health/ammo, rapidly turning around to face enemies immediately after finding ammo (see Figure \ref{fig:orge599d98}) and human like navigation around buildings. All these behaviours can be more fully appreciated in the video.

% An example of a double column floating figure using two subfigures.
% (The subfig.sty package must be loaded for this to work.)
% The subfigure \label commands are set within each subfloat command,
% and the \label for the overall figure must come after \caption.
% \hfil is used as a separator to get equal spacing.
% Watch out that the combined width of all the subfigures on a 
% line do not exceed the text width or a line break will occur.
%
%\begin{figure*}[!t]
%\centering
%\subfloat[Case I]{\includegraphics[width=2.5in]{box}%
%\hfil
%\subfloat[Case II]{\includegraphics[width=2.5in]{box}%
%\label{fig_second_case}}
%\caption{Simulation results for the network.}
%\label{fig_sim}
%\end{figure*}
%

\section{Future Work}
In future work we aim to further enhance the capabilities of the MAIL architecture by adding continuous actions for all rotations. This should provide a number of benefits when combined with the current MAIL architecture. Not only will it provide the agent with more fine grained motor control and reduce the size of the action space, it will also allow much higher quality expert data to be recorded by allowing data to be acquired using a mouse and keyboard or the analogue inputs on a game controller. These improvements should allow the MAIL architecture to be used to train agents to play modern AAA FPS games. Relaxing assumption \ref{mainassumption} to more general forms of parametric policies is also left for future work.

% or


%\appendices

%\section{}

\section*{Acknowledgments}
We would like to thank Paul Greveson and Ken Brown for help with game art, Dirk de la Hunt for help with game engine technology and Martin Singh-Blom for insightful discussions.

  %% \newpage



%\bibliographystyle{IEEEtran}

\bibliographystyle{abbrvnat}
\bibliography{bib}

%% \begin{thebibliography}{1}
%% \bibitem{IEEEhowto:kopka}
%% H.~Kopka and P.~W. Daly, \emph{A Guide to \LaTeX}, 3rd~ed.\hskip 1em plus
%%   0.5em minus 0.4em\relax Harlow, England: Addison-Wesley, 1999.
%% \end{thebibliography}

\end{document}